# Relative Distance Features for Gait Recognition with Kinect

Ke Yang [a, b*], Yong Dou[a, b], Shaohe Lv[a, b], Fei Zhang[a, b], Qi Lv[a, b]

[a] National Laboratory for Parallel and Distributed Processing,

National University of Defense Technology, Changsha 410073, China

[b] College of Computer, National University of Defense Technology,

Changsha 410073, China

## Abstract

Gait and static body measurement are important biometric technologies for passive human recognition. Many previous works argue that recognition performance based completely on the gait feature is limited. The reason for this limited performance remains unclear. This study focuses on human recognition with gait feature obtained by Kinect and shows that gait feature can effectively distinguish from different human beings through a novel representation -- relative distance-based gait features. Experimental results show that the recognition accuracy with relative distance features reaches up to 85%, which is comparable with that of anthropometric features. The combination of relative distance features and anthropometric features can provide an accuracy of more than 95%. Results indicate that the relative distance feature is quite effective and worthy of further study in more general scenarios (e.g., without Kinect).

---

[*] Corresponding author. E-mail address: bityangke@163.com (Ke Yang)

## 1. Introduction and Related Works

Accurate and effective human recognition is a major research area of computer vision, pattern recognition, machine learning, biometrics, and intelligent surveillance. Face [1] and fingerprint recognition [2] technologies have been widely used in commercial and forensic fields. However, such biometrics need to be detected with subjects' cooperation and require high image resolution. A relatively recent trend in biometrics is performing person recognition using full-body characteristics [9]. In contrast to traditional identification technologies, human recognition technology using full-body characteristics can be used from a distance and does not need subjects' cooperation. A large number of surveillance systems built to improve safety and security have been developed recently. Many of these systems include human recognition capabilities, and recognition method using full-body characteristics may be the most suitable one. A surveillance video mainly aims to monitor people; however, most surveillance cameras can only collect low quality videos, and traditional biometrics may be concealed. Low quality videos can still provide sufficient data for the full-body recognition technology, and studies using this biometrics as a forensic tool exist [19]. Full-body characteristics include static body measurement and gait features [9, 17].

Static body measurement refers to static body shape [17] or anthropometric feature [9]. Araujo et al. [14] extracted 11 anthropometric features, including skeleton length and height to recognize different people. Gait features model the walking pattern of people. Existing gait recognition methods can mainly be divided into two groups: model-free and model-based technologies [20]. Model-free gait recognition technologies include a variety of silhouette-based methods. Phillips et al. [18] proposed a baseline algorithm by using the correlation of silhouettes. Gait Energy Image (GEI) [3] uses the average image of silhouettes in a gait period to characterize gait features. GEI has become one of the most common methods in recent years because of its simple representation and computation effectiveness. However, GEI's performance could be affected by silhouette quality, namely, segmentation errors. According to reference [3], GEI is insensitive to silhouette segmentation errors. But in the reference [20], the author argue that it is only true for pixel-wise independent random noise where false and miss detection of foreground are with the same probability. When it comes to non-random noise where background and foreground are similar, the segmentation will always be wrong, so GEI is significantly affected by the silhouette quality. An enhanced version of GEI, namely, alpha-GEI is proposed to mitigate such a non-random noise in reference [23]. The reference [24] improves the silhouette quality and recognition accuracy by using standard gait models as prior knowledge. Reference [25] extends the quality metrics and proposes a novel way of using quality metrics to improve the segmentation pre-processing step. Gait Entropy Image (GEnI) [4] and Chrono-Gait Image (CGI) [5] are also silhouette-based gait representations proposed based on GEI [3]. GEnI [4] mainly has encoding dynamic gait features and is robust to variable factors, such as clothes and backpacks. CGI [5] uses color to encode gait features to store temporal information. Model-free gait recognition method possesses simple representations, extractable features, and low computational complexities, but addresses viewpoint and occlusion problems poorly [20]. Therefore, there are plenty of works which enhanced the model-free method to handle cross-view [26, 27, 28, 29] and occlusion problems [30, 31]. Model-based gait recognition methods model the structure of the human body by using body structure parameters [20]. Cunado et al. [6] developed an early model-based method called the pendulum model. The thigh modeled as linked pendulum and gait features are extracted from the frequency component of inclination angle signals. Stick model [7] extracts head, neck, shoulder, chest, pelvic, knee, and ankle positional parameters in the body silhouette according to the knowledge of anatomy and then calculates the kinematic characteristics of various locations to construct the 2D human stick model. All stick models in one gait sequence are linked together to form the gait pattern for recognition.

Many previous studies argue that recognition performance based completely on the gait feature is limited [9, 16, 17]. Lombardi et al. [17] proposed a two-point gait that encodes a complete human dynamic gait feature and found that recognition ability using a two-point gait separately is limited, but when combined with the body shape features, the recognition accuracy has a significant improvement. The experimental results of Veeraraghavan et al. [16] showed that body shape is more important than gait features in a video-based gait recognition task. Andersson et al. [9] also suggested that using static anthropometric features performs far better than using gait features in skeleton-based human recognition. The present study aims to determine whether the distinguishing ability of the gait itself is limited, or existing features do not adequately capture the right features. This study investigates this problem in Kinect-based human recognition.

Model-based gait recognition methods are robust to occlusion and viewpoint but require complex calculations to extract the human body model. Thus, most of the works are based on model-free methods. The emergence of Kinect has changed this phenomenon [8]. Kinect is the motion controller of XBOX-360 released in 2010 by Microsoft. Conventional color image, depth image of the scene, and human skeleton data stream can be simultaneously extracted from Kinect. The skeletal tracking function of Kinect can provide real-time 3D coordinates of 20 human skeleton points, eliminating the need for complex extraction procedures of human model and providing a great convenience for robust gait recognition technology to light, viewpoint. Similarly, full-body characteristics used in Kinect skeleton-based human recognition also include static body measurement and gait features [9].

The most common division between gait features, according to the human gait theory, is temporal parameters, spatial parameters and kinematic parameters [21]. Spatial and temporal parameters are the intuitive gait features including step length, speed, gait cycle, average stride length, and so on. Kinematic parameters are usually characterized by the joint angles between body segments and their relationships to the events of the gait cycle [21]. Preis et al. [8] extracted spatiotemporal parameters – step length and speed to perform the human recognition task. Since both of these features are changing constantly during walking, using these features is easy to confuse different people. Andersson et al. [9] extracted the temporal parameters, spatial parameters and kinematic parameters such as step length, cycle time, speed and features extracted from lower limb angles as gait features. Angle-based kinematic features have poor performance because lower limb angles are disturbed by noise. When people walk in parallel to Kinect, the joints of the distant body parts may be occluded and inferred; when the joints used are inferred, extracting kinematic features from lower limb angles as gait features is inaccurate. And the references such as [9] use the angle-based features from both sides. Ball et al. [10] also extracted the lower limb angle-based kinematic features as gait features.

In several works such as [11, 12], distance-based gait features are proposed to model the human gait. The statistical characteristics of the distance of one sequence such as mean values are extracted as the gait features, and the distance-based gait features have distinguishing ability to some extent. And the distance-based gait features can be divided into two categories: relative distance-based and absolute distance-based gait features. These two classes of features can also name them as relative distance features and absolute distance features. In the Kinect depth sensor coordinate system [Fig. 1(a)], absolute gait features are the statistical characteristics extracted from the relative coordinate values of specific skeleton point pairs.

Vertical Distance Features (VDF) was developed by Ahmed et al. in [11]. VDF describes the statistical characteristics of the absolute coordinate value changes of some joints when walking. In the most common scenarios that people walk parallel to the Kinect depth sensor's coordinate system's x-axis, the curve of skeleton points' absolute coordinate values (in the coordinate system in Fig. 1(a), it represents the x-axis direction) is not periodic, and the curve can be approximately modeled by an $n$-order trigonometric-polynomial interpolant function according to reference [21]. The curves lose the periodic characteristic of the gait and the parameters of the function need a relative long sequence to fit. The curve of the relative coordinate values of specific skeleton point pairs is still periodic, keeping the periodic characteristic of gait. Thus, the relative distance features are more suitable to model the gait from this perspective.

Gianaria et al. [12] extracted spatiotemporal parameters such as step length, speed, relative distance features (such as average distances between two elbows, two hands, and two ankles), and also the absolute gait features extracted from some joints' absolute coordinate changes as gait features. Chattopadhyay et al. [13] used the relative distance features (such as the mean values of relative distance from front elbow, wrist, hand, ankle, and foot to hip center) and hip center's velocity as gait features. Using statistical characteristics of the relative motion of unconnected joints as representation of gait is robust; however, only using the mean value is insufficient. The standard deviation of the relative distances models the amplitude of movement of people when walking. Since the amplitudes of different people' skeleton joints' movements are different, the standard deviation has the discrimination ability to some extent. It could be a feasible way to combine the mean value and the standard deviation to achieve better performance.

In this study, we extract relative distance features (i.e., the mean and standard deviation of relative coordinate values between the joints) as gait features. The recognition accuracy of the proposed robust relative distance features is up to 85%. This accuracy result is comparable with the anthropometric features and does not require calculation of the gait cycle. Moreover, relative distance features and anthropometric features complement each other well. Recognition accuracy of more than 95% can be achieved when these features are combined. The experimental results have verified that gait features own sufficient recognition capability; relative distance features are an effective representation of gait in the Kinect-based scenario, and worthy of further study in general scenarios.

The rest of the paper is organized as follows. The proposed method is presented in Section 2. The experiments and results are described in Section 3. Conclusion is presented in Section 4.

## 2. Proposed Method

This section describes the proposed method and is divided into five parts: Kinect skeleton data stream, the extraction of gait features, the anthropometric features, feature selection and classification.

As mentioned in the Section 1, we use the similar anthropometric features as [9, 14]. We also use the similar relative distance features as [13], but we choose different relative distance set and add the standard deviation value as features to improve the gait features' distinguishing ability. Based on the extracted features, we apply the random subspace method and K-nearest neighbor classifier to accomplish the feature selection and classification process. We have evaluated our method on the newly released 140-subjects Kinect skeleton dataset [9].

## 2.1 Kinect Skeleton Data

Kinect is the motion controller launched by Microsoft in 2010 and was an XBOX-360 accessory initially. Researchers have been using Kinect to develop various applications because it can capture RGB images, depth images of a scene, skeleton data streams, and voice data streams in real time.

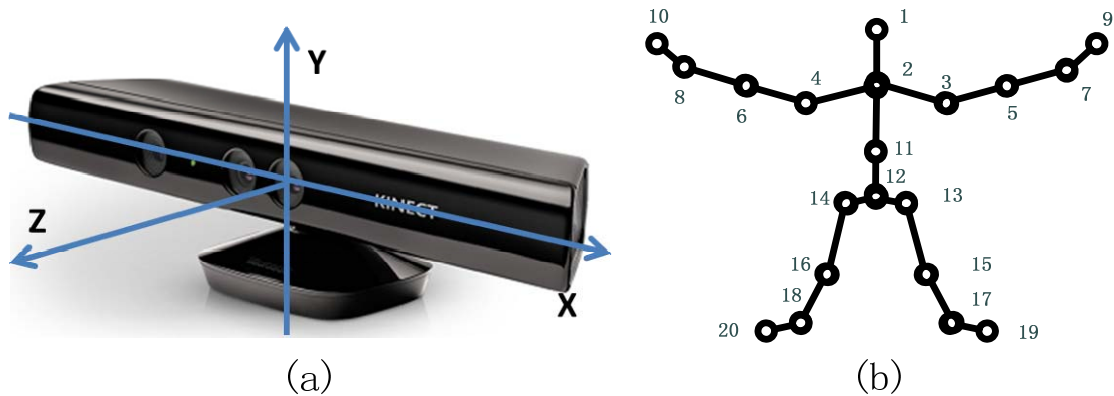

(a)                          (b)

**Fig. 1.** (a) Kinect depth sensor coordinate system (b) Twenty joints of skeleton.

The proposed method only uses skeleton data stream. Kinect v1.0 can provide 3D coordinates of the 20 joints of two human bodies at 30 fps [Fig. 1(b)]. The names of each joint are shown in Table 1. The origin of the Cartesian coordinate system is the depth sensor's center; $x$-axis is parallel with Kinect; $y$-axis is vertical to the bottom surface of Kinect (if Kinect is settled horizontally, it is the horizontal plane); and $z$-axis is the direction parallel to the sensor's normal direction [Fig. 1(a)]. The units are in meters. This paper extracts dynamic and anthropometric features based on Kinect skeleton data stream to identify people.

**Table 1**

Skeleton joints names.

| Joint Num. | Joint Name | Joint Num. | Joint Name |
|---|---|---|---|
| 1 | Head | 11 | Spine |
| 2 | Shoulder-Center | 12 | Hip-center |
| 3 | Shoulder-Right | 13 | Hip-Right |
| 4 | Shoulder-Left | 14 | Hip-Left |
| 5 | Elbow-Right | 15 | Knee-Right |
| 6 | Elbow-Left | 16 | Knee-Left |
| 7 | Wrist-Right | 17 | Ankle-Right |
| 8 | Wrist-Left | 18 | Ankle-Left |
| 9 | Hand-Right | 19 | Foot-Right |
| 10 | Hand-Left | 20 | Foot-Left |

2.2 Relative Distance Based Gait Feature Extraction

As mentioned in the Introduction and Related Works Section, the curves of the absolute coordinates may lose the periodic characteristic of the gait and the parameters of the function need a relative long sequence to fit. The curves of relative coordinate values of specific skeleton point pairs are still periodic curves, maintaining the periodic characteristic of gait. Thus, it is more reasonable to use relative distance features to model the gait. All gait features are selected based on the relative coordinates of the joints (i.e., relative distance features). The extracted gait features are divided into three directions: $x$, $y$ and $z$. The directions are identical to the coordinate system's axis direction. The features of $x$ (or $y$, $z$) direction mainly refer to the statistical characteristics (mean values and standard deviations) of relative distances of some joints in the $x$ (or $y$, $z$) direction. The 11 relative distances are shown as Equation (1):

$$Dx1 = abs(x(17) - x(18)),$$
$$Dx2 = abs(x(5) - x(6)),$$
$$Dx3 = abs(x(9) - x(10)),$$
$$Dx4 = abs(x(1) - (x(17) + x(18)) / 2),$$
$$Dx5 = abs(x(11) - (x(17) + x(18)) / 2),$$
$$Dx6 = abs(x(7) - x(8)), \tag{1}$$
$$Dx7 = abs(x(3) - x(4)),$$
$$Dy1 = abs(y(1) - (y(19) + y(20)) / 2),$$
$$Dy2 = abs(y(1) - (y(15) + y(16)) / 2),$$
$$Dy3 = abs(y(19) - y(20)),$$
$$Dz1 = abs(z(9) - z(10)).$$

where ($Dxi$ or $Dyj$, $Dzk$) represents the distances in $x$-axis (or $y$-axis, $z$-axis) direction and $x(i)$ (or $y(j)$, $z(k)$) represents the $i$-th (or $j$-th, $k$-th) joint's $x$ (or $y$, $z$) coordinate value. The $abs(\cdot)$ stands for the absolute value function. For example, $Dx1$ is the distance between two ankles in the $x$-direction.

The aforementioned 11 relative distances can derive the following features:

$$MEAN = mean\{Dx1, Dx2, Dx3, Dx4, Dx5, Dx6, Dy1, Dy2, Dy3, Dz1\},$$
$$STD = std\{Dx1, Dx2, Dx3, Dx4, Dx5, Dx6, Dx7, Dy1, Dy2, Dy3\}, \tag{2}$$
$$RDF = \{MEAN, STD\}.$$

where $mean(\cdot)$ is the mean value function and $std(\cdot)$ is the standard deviation function. $MEAN$ is the vector of mean value of (Dx1, Dx2, Dx3, Dx4, Dx5, Dx6, Dy1, Dy2, Dy3, Dz1), and $STD$ is the vector of standard deviation of (Dx1, Dx2, Dx3, Dx4, Dx5, Dx6, Dx7, Dy1, Dy2, Dy3). The two sets are combined into a relative distance features vector $RDF$, whose dimension is 20.

To select the relative distances introduced in Equation (1), we carry out the following steps: first, we list almost all of possible relative distances in three axis directions; then we calculate the MEAN and STD vector of all the relative distances to get the RDF; next, we carry out the greedy procedure -- forward selection [35] to choose the raw features combination to perform the feature selection with the random subspace method. The relative distances introduced in Equation (1) can be obtained from the raw features combination easily.

2.3 Anthropometric Feature Extraction

Similar to other studies [9, 14], the proposed anthropometric features mainly refer to the length of different body parts. All the selected 19 segment lengths set *Skeleton Length* [Equation (3)] and *Height* [Equation (4)] are combined to create an anthropometric features set *AF* [Equation (5)], whose dimension is 20. The height is defined as the sum of neck length, upper and lower spine length, and average leg length. In contrast to studies [9, 14], height calculation does not use left and right hip lengths.

$$Skeleton\ Length = \{Distance_{Euclid}(i,j) \mid joint\ i\ and\ joint\ j\ are\ connected\}, \qquad \textbf{(3)}$$

$$Height = Distance_{Euclid}(1,2) + Distance_{Euclid}(2,11) + Distance_{Euclid}(11,12) + \qquad \textbf{(4)}$$
$$(Distance_{Euclid}(14,16) + Distance_{Euclid}(16,18) + Distance_{Euclid}(13,15) + Distance_{Euclid}(15,17))/2,$$

$$AF = \{Skeleton\ Length, Height\}. \qquad \textbf{(5)}$$

where $Distance_{Euclid}(i,j)$ represents the Euclidean distance between the *i*-th and *j*-th joints.

*AF* is calculated for each frame. The mean and standard deviation are calculated over all the past frames. The mean of each component is recalculated after removing outliers that are over two standard deviations from the mean. Then the new mean makes up the final *AF* vector.

The relative distance feature vector *RDF* proposed in Section 2.3 and the proposed *AF* comprise the combined feature vector *CF*, whose dimension is 40. To sum up, this study propose three feature sets, namely, *RDF*, *AF*, and *CF*={*AF*, *RDF*}.

2.4 Feature Selection

Numerous noisy data exist when Kinect skeleton data stream is captured, resulting in the destruction of certain features. On one hand, using bone length as a feature is inappropriate if the joint is occluded and inferred. On the other hand, what features are relevant is uncertain because of unknown walking conditions. Thus, a fixed feature selection is difficult. Inspired by [15], a classifier ensemble method based on Random subspace method (RSM) and majority voting(MV) is employed as feature selection and classification method to track this problem. To the best of our knowledge, this is the first time that feature selection method is used in Kinect-based gait recognition. As shown in Fig. 2(a), $F$ (i.e., $RDF$, $AF$, and $CF$) represents the feature space; $R_F^i$ stands for $i_{th}$ ($i = 1, ..., L$) random subspace of $F$; and $R_F^i$'s dimension is N = 10, that is to say N = 10 features are randomly selected from $F$ for each weak classifier. The classification results are obtained from the voting results of L = 100 weak classifiers. The RSM method is summarized in Algorithm 1. Our focus is to validate the effectiveness of RSM in this scenario, thus we do not pay much attention to parameter adjustment, and we set these parameters empirically.

**Algorithm 1 RSM-MV with KNN**

**Input:** Feature set (i.e., the combined feature vector CF) $F \in R^{1 \times N_1}$, weak classifiers number $L$, the dimension of the subspace $N$;

**Output:** The classification result $LABEL$;

**Step 1:** Generate the random permutation index vector

$r^i = randomsample(\{1, 2, ..., N\})$, $i \in [1, L]$;

**Step 2:** Construct the $L$ feature subspaces:

Feature selection on Feature set (Extract the corresponding rows' feature)

$R_F^i = F(p^i)$, $i \in [1, L]$;

**Step 3:** Use each weak classifier to get the labels of the $L$ subspaces

$label_i = WeakClassifier(R_F^i)$, $i \in [1, L]$;

**Step 4:** Output the ensemble classification result $LABEL$ through MV

$LABEL = MajorityVoting(\{label_1, label_2, ..., label_L\})$.

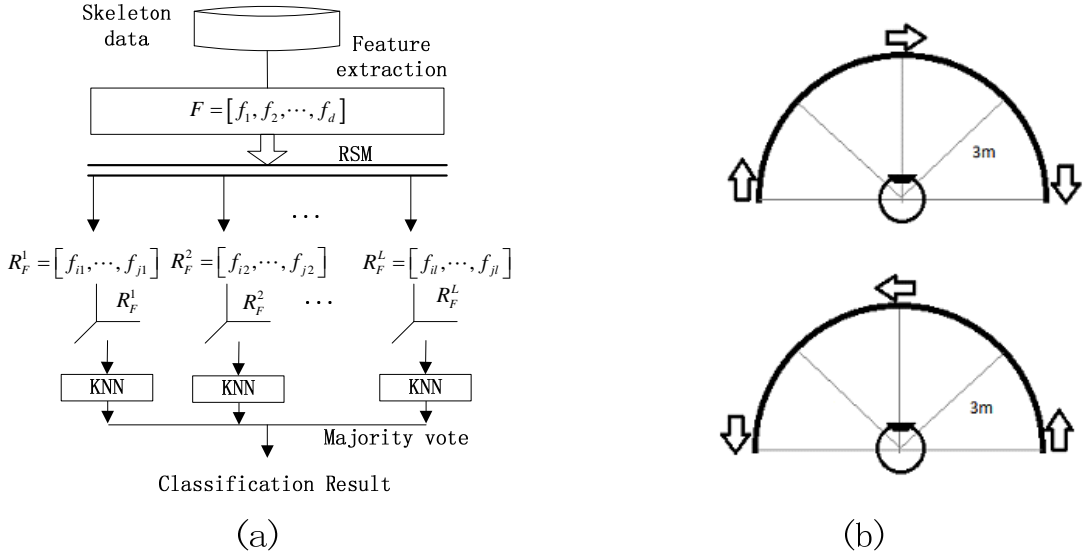

(a)                                                    (b)

**Fig. 2.** (a) Feature selection and classification and (b) semi-circular path used for subjects' walks. The Kinect sensor at the center is equipped with a dish to allow for tracking [9].

## 2.5 Classifier

Many researchers have shown that K-nearest neighbor (KNN) and support vector machines are the best choice for classifiers in gait recognition task [9, 14, 21]. In this study, the KNN classifier with Manhattan distance as the distance metric is employed for classification. Given a probe sequence $p$, first we compute the Feature $F_p$, then compute distances with $F_g{}^i \in R^{1 \times N_1} (i = 1, ..., n_g)$ [Equation (6)], $n_g$ is gallery size; finally, find the K "closest" labeled examples in the gallery and assign $p$ to the class that appears most frequently within the K-subset.

$$Dis \tan ce(F_p, F_g^i) = \sum_{j=1}^{N_1} | f_p^j - f_g^{ij} |, \ i \in [1, ..., n_g].$$ **(6)**

## 3. Experiments and Results

### 3.1 Skeleton Gait Dataset

The public skeleton gait dataset proposed by Andersson et al. [9] is used for this study. The dataset includes 140 subjects, and each has five sequences. In the data capturing procedure, each volunteer was asked to walk in front of the Kinect in a semi-circle. A spinning dish helps Kinect to keep the subject always in the center of its view. Each subject executed five round-trip free cadence walks, starting from left, walking clockwise to the right, and then back [Fig. 2(b)]. Each sequence includes about 500 to 600 frame data.

3.2 Classification Accuracy

In our method, KNN is chosen as classifier like most of the gait recognition algorithms. When using the KNN classifier, the accuracy varies with the value of K. Usually, when the data is suffered with noise, the accuracy of KNN classifier will be low when K is given a small value because the nearest neighbor might be noisy data. For most studies, the optimal value of K is always obtained through experiment. In the gait recognition task, KNN classifier is widely used. In reference [15], 1-NN (i.e., K is equal to 1) is used as classifier. In reference [21], the authors use KNN as classifier, and their experiment results show that the performance is best when K is equal to 1, so as the references [14, 15, 32].

In this study, 10-fold cross-validation is used to select the optimal K value. The dataset is randomly partitioned in 10 subsets, and training is performed 10 times, each subset leaves one partition out of training process for testing. The accuracies are the average of the 10 executions. The search range of K is from 1 to 70. The results are shown in Fig. 3. As the figure shown, the accuracy of all three feature sets reaches the highest value when K = 1. Thus, all subsequent experiments use the parameter K = 1 (i.e., k-NN becomes 1-NN).

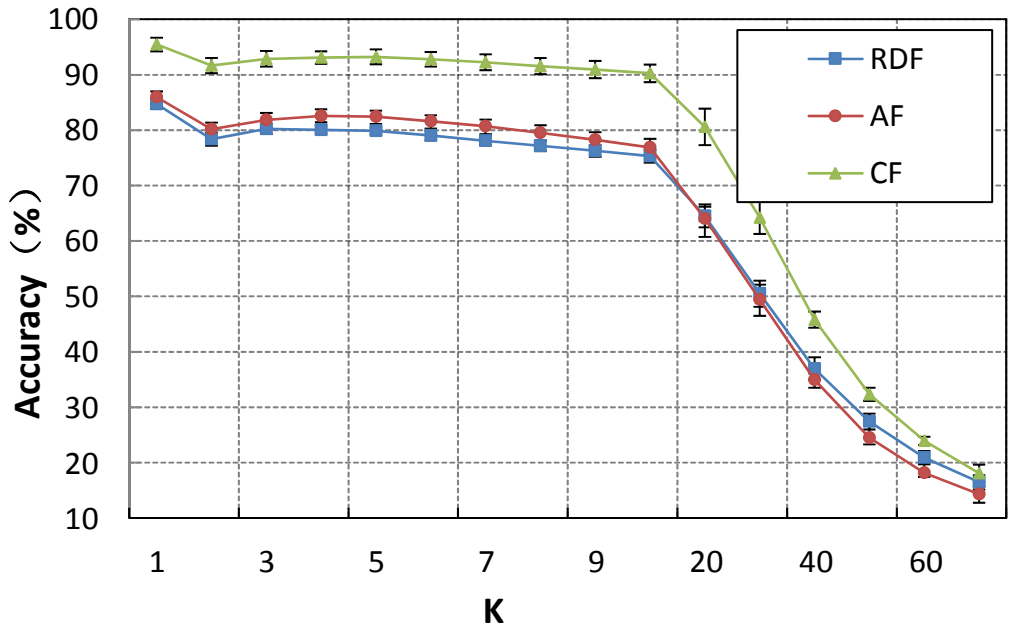

**Fig. 3.** Average classification accuracy of 10-fold cross-validation for different K values.

Table 2 shows the average recognition accuracies when using different sets of features. The relative distance features' accuracy of 140 subjects is 84.6%, which is comparable with the anthropometric features. When relative distance features and anthropometric features are combined, 95.4% of recognition accuracy can be achieved, with a 9.3% improment compared with anthropometric features. This result suggests that the proposed relative distance features and anthropometric features complement well.

**Table 2**

Average recognition accuracies under different feature sets when the gallery size is equal to 140

| Feature set | AF | RDF | CF |
|---|---|---|---|
| Accuracy | 86.1% | 84.6% | 95.4% |

Given the lack of large public dataset, researchers used their own captured dataset with various gallery sizes. Different dataset sizes may make identification with different difficulty level. The recognition accuracy can still be high if a feature set's performance is poor but the gallery size is small. Thus, similar to [9] the proposed method was evaluated under different gallery sizes. For each size P (P = 10, 20…, 30), 10 subsets of the same size were randomly drawn for gallery and the accuracies are the average of 10 subsets. The results are shown in Fig. 4. The proposed three feature sets' recognition accuracies are not very sensitive to changes in size, especially when using combined features. The experimental results on various gallery sizes further validated that gait features have a sufficient recognition capability that is comparable to anthropometric features.

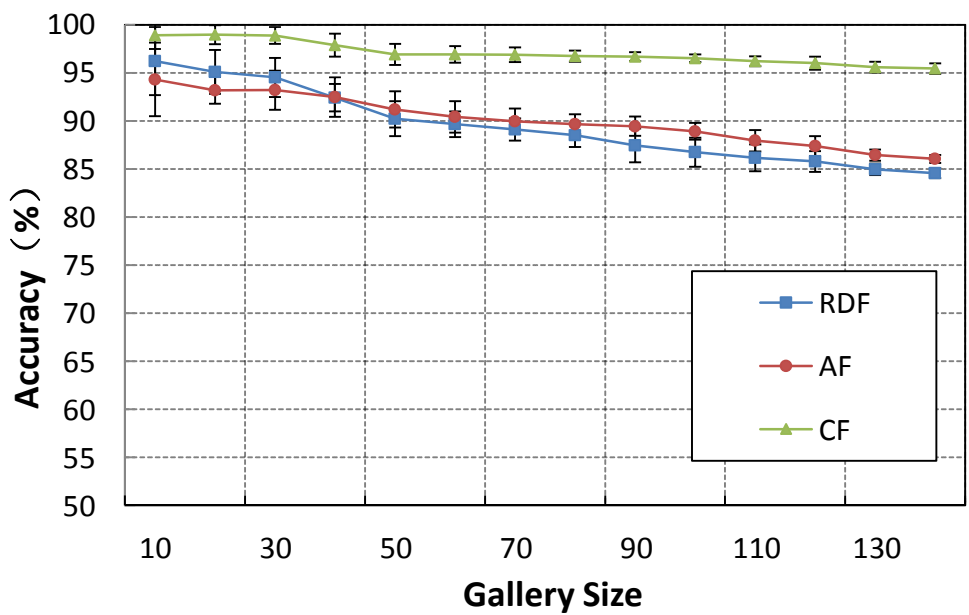

**Fig. 4.** Average recognition accuracy under different gallery sizes and different feature sets (error bars represent one standard deviation).

People are more concerned about a recommended sequence than the best match in several classification tasks. When solving cases using biometrics, the police care about a recommended sequence of a few people rather than the most suspicious person. Thus, cumulative match curves (CMCs) is also an important evaluation method of performance. CMCs of three proposed feature sets are shown in Fig. 5. Rank-5 accuracy that uses combined features is up by 99%, which is almost enough for practical use.

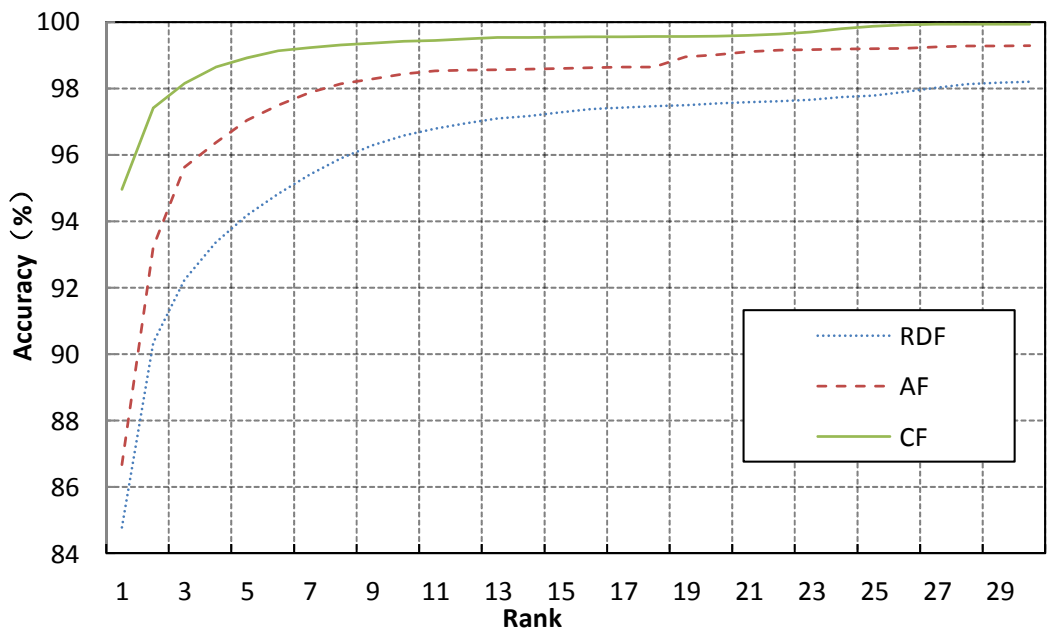

**Fig. 5.** CMCs when using the proposed three feature sets.

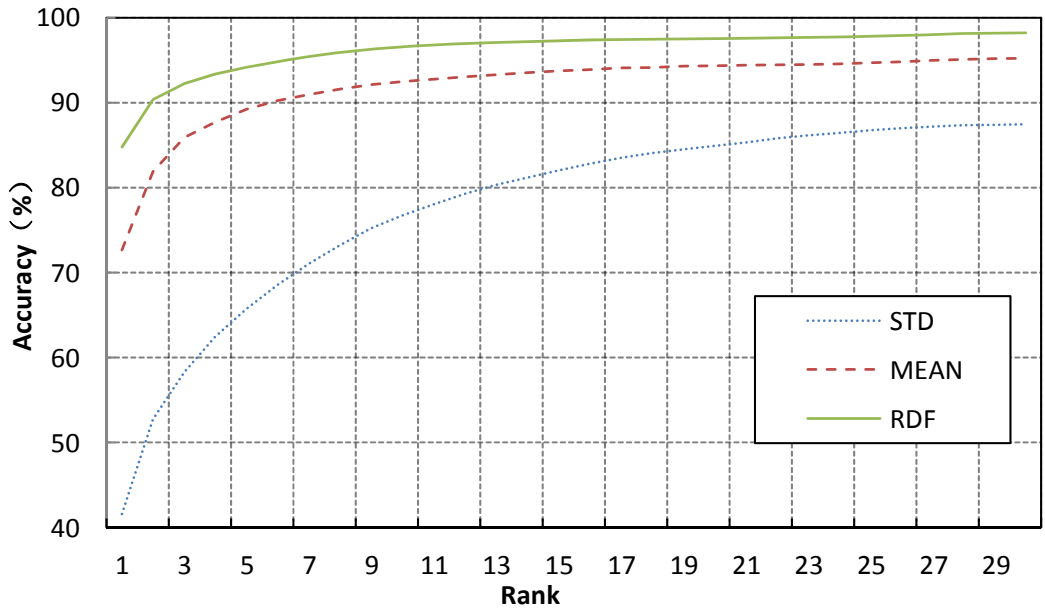

**Fig. 6.** CMCs when using the proposed relative distance feature and its subsets.

When introducing the relative distance features of skeleton-based human recognition, it is mentioned that using only the means of relative distances is not enough and the standard deviation also can identify people to some extent. The combination of the two may lead to a better recognition performance. To investigate this, *RDF* 's subsets *MEAN* and *STD* are separately used for recognition to prove this viewpoint. The results are reported in Fig. 6 using CMCs. Separately employing *MEAN* 's performance provides unsatisfactory results, but the performance improves greatly when *STD* is combined with *MEAN* , confirming the previous analysis.

3.3 Performance Improvement of Random Subspace Method

The proposed method used RSM for feature selection on the proposed feature space. The percentage of improvement RSM brings is yet to be determined. This study also conducted experiments without feature selections and classifier ensemble. The results are shown in Table 3 and Fig. 7. When using all the 140 subjects, RSM exhibits accuracy improvement of 2.2%, 0.8%, and 1.0% for anthropometric features, relative distance features, and combined features, respectively (Table 3). Figure 7 shows how the accuracies vary in both with and without RSM for different gallery sizes. Performance improvement is not sensitive to different gallery sizes. Taking account of only using empirical parameter settings, it is reasonable that RSM does not cause a large increase in recognition rates.

**Table 3**

Classification accuracy using three feature sets when using all the 140 subjects

| Feature Selection | AF | RDF | CF |
|---|---|---|---|
| Without RSM | 83.9% | 83.8% | 94.4% |
| With RSM | 86.1% | 84.6% | 95.4% |
| Improvement | 2.2% | 0.8% | 1.0% |

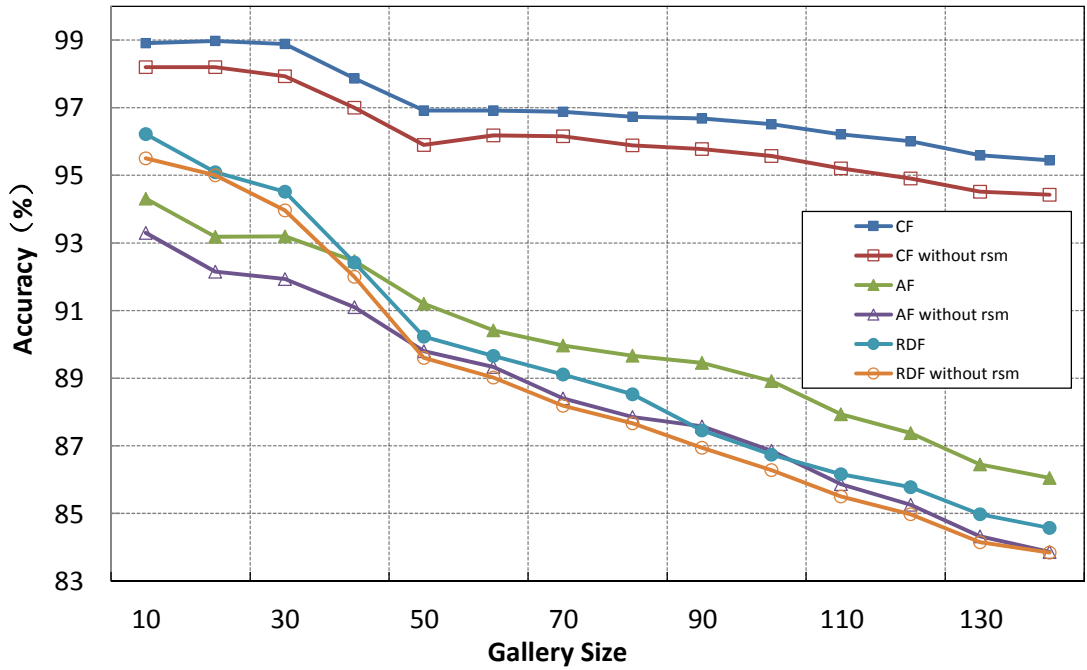

**Fig. 7.** Classification accuracy under different feature sets and different gallery sizes.

3.4 Comparisons with Other Methods

This study mainly investigates the gait features with strong recognition capacity to characterize the kinetic characteristic of human gait. Thus, comparisons mostly occur in different gait features. The development of Kinect-based human recognition has not been perfect. Its imperfection significantly manifests in the lack of convictive public dataset. Almost all studies were conducted on their own established datasets, leading to the difficulty in comparing different methods. Moreover, given the differences in preprocessing methods, the distance between subjects and Kinect, walking directions, and classifiers used, a convictive and quantitative comparison is almost impossible to achieve. Because all the results of related works use their own dataset, so the comparisons parts in the related works are mainly qualitative comparisons [9, 13]. To give an intuitive and qualitative comparison, 10 same-sized subsets with each related work of dataset from [9] are randomly selected, and the average accuracies are calculated on the 10 random subsets. Given that the related studies may have used more than one classifier or a set of features, only the highest results of their works are reported in Table 4. Note that, results in Table 4 use different datasets except study [9] with our method, thus Table 4 just gives a qualitative comparison. The dataset used in our method is the dataset from the state-of-the-art work [9]. To give a convincing quantitative comparison, and considering the development period, we have implemented the work in [11] because it "seems" to have the best results. We compare our method with [9, 11] and present the results in Fig. 8, using the same dataset captured in [9].

**Table 4**

Qualitative comparisons with other methods

| Related work | No. of Subjects | Accuracy of Dynamic Features | Our Method |
|---|---|---|---|
| Ball et al. [10] | 4 | 43.6% | 98.8% |
| Preis et al. [8] | 9 | 55.2% | 95.9% |
| Gianaria et al. [12] | 20 | 52% | 95.1% |
| Gianaria et al. [33] | 20 | 96% | 95.1% |
| Ahmed et al. [11] | 20 | 83.5% (VDF) | 95.1% |
| Chattopadhyay et al. [34] | 29 | 75% | 94.6% |
| Chattopadhyay et al. [13] | 60 | 55.26% | 89.7% |
| Andersson et al. [9] | 140 | 62.9% | 84.6% |

According to the qualitative comparison results in Table 4, the proposed relative distance features provide comparable performance with [33] and outperform other methods. The recognition accuracies of spatiotemporal parameters [8] and angle-based kinematic parameters [9, 10] are very low, confirming the analysis in the related works section. Besides, in any case the results of present study are good considering the huge gallery size (140 subjects), much bigger than the ones used in other works except [9]. Figure 8 shows the variation of recognition accuracies of [9, 11] and the proposed method under different gallery sizes.

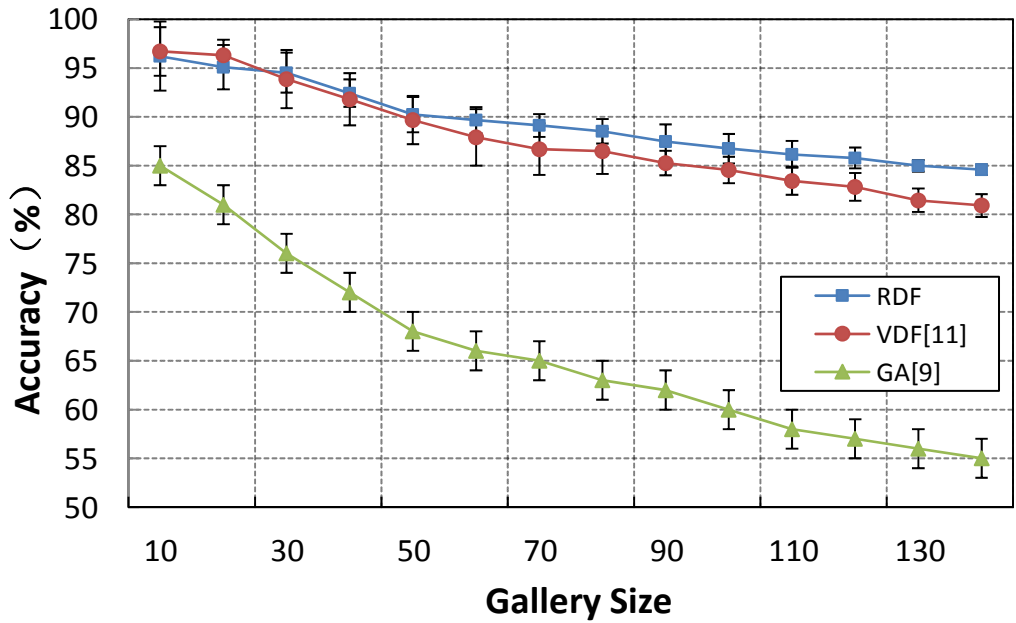

**Fig. 8.** VDF [11], GA [9], and the proposed relative distance features' average recognition accuracy under different gallery sizes and feature sets. Error bars represent one standard deviation.

As shown in Fig. 8, the recognition accuracy of gait features proposed in [9] (i.e., GA) decreases rapidly with increasing gallery size according to the results reported therein. The VDF used in Fig. 8 is implemented by us and experimented on the 140-subject dataset. The VDF's recognition accuracy is slightly higher than the proposed method when the gallery size is very small, such as 20. However, error bars indicate the difference is not significant. The recognition accuracy of implemented VDF decreases more rapidly than the proposed relative distance features as gallery size increases. The performance of the proposed relative distance features is significantly better than the implemented VDF's when the gallery size is larger than about 20. Therefore, the discrimination of the relative distance gait features is higher than angle-based kinematic parameters and absolute distance features.

Moreover, recognition accuracy of the proposed relative distance features reaches up to 95.4% when combined with the proposed anthropometric features; this result is better than the state-of-the-art technology [9] by 7.7%.

The response time of an algorithm is important for judging the practical applicability, so we test the response time of our method. Our method took about 0.04 seconds to process a sequence on a desktop machine with Intel i5 CPU using a single thread. The result shows that our method is quite efficient and is with high practical applicability.

3.5 Discussion

The data acquired with Kinect sensor v1 is truly subject to acquisition errors. Considering the demand for online identification, the proposed method does not filter raw skeleton data. In [15], Guan et al. have verified that the random subspace method is robust to the occupied features, even for data with acquisition errors. Hence, we used the random subspace method to select the features.

The discrimination of the relative distance features is higher than that of angle-based kinematic parameters and absolute distance features because the latter features are more affected by the joint occlusion problem. Hip and Knee angles [9], knee absolute coordinates [12], and the proposed relative distance features are extracted and evaluated to give an intuitive visual comparison (Fig. 9 and Fig. 10). Solid line curves in Figs. 9 (a), (b), and (c) represent the normal and perfect collection by reference to curves in [7]. Dotted line and dash-dot line curves represent the actual and noisy collection, respectively. These curves are extracted from the 140-subject public dataset [9].

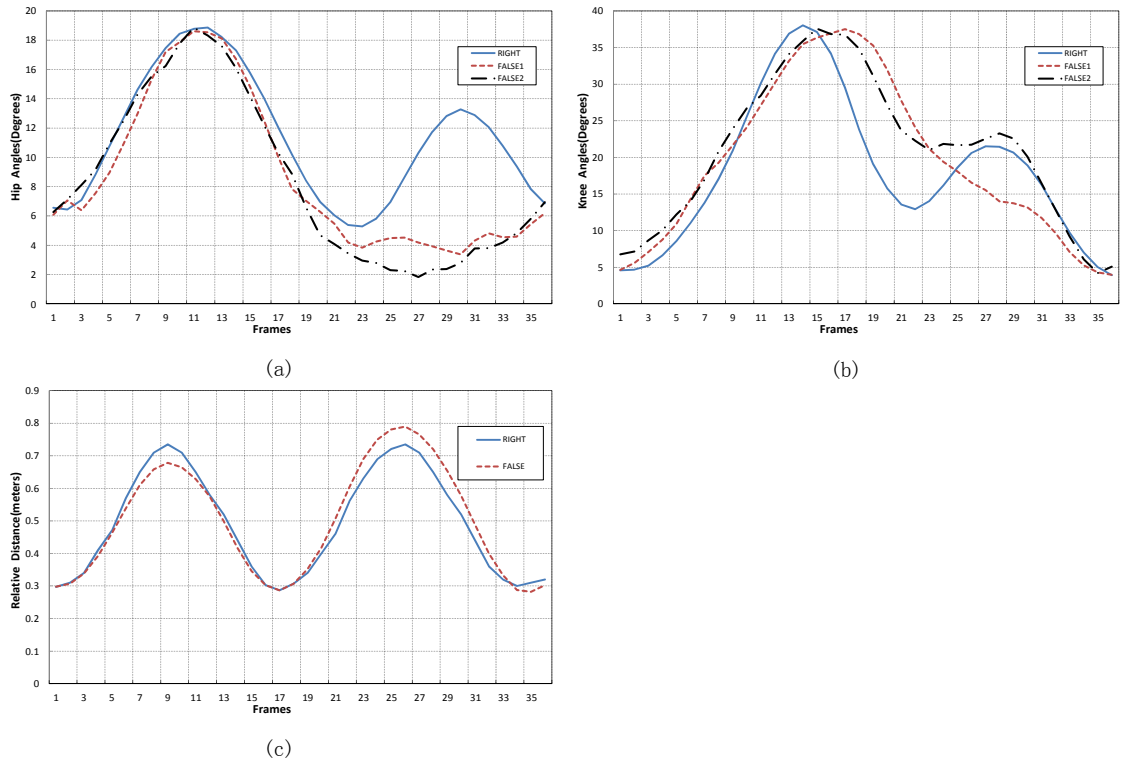

**Fig. 9.** (a) Hip angles, in degrees, from the right leg, corresponding to a gait cycle (b) knee angles, in degrees, from right leg, corresponding to a gait cycle (c) relative distances between two ankles, in meters, corresponding to a gait cycle; 'RIGHT' represent the clean data, 'FALSE' represents noisy data.

The absence of one peak in some gait cycles occurs to hip angles [Fig. 9(a)]. Knee angle curves in most gait cycles have lost one peak indicated by the red dotted line in Fig. 9(b) and in some gait cycles lost the valley between two peaks indicated by the black dash-dot line. Proposed relative distance feature curves, such as the relative distance between two ankles, are shown in Fig. 9(c). Only the amplitudes are unstable, but the trend is right. Thus, relative distance features are less affected by noisy skeleton data than angle-based kinematic features. This result is because the distance calculation only has one occluded joint. However, angle calculation has more than one occluded joint, resulting in the rotation of bones being more uncontrollable.

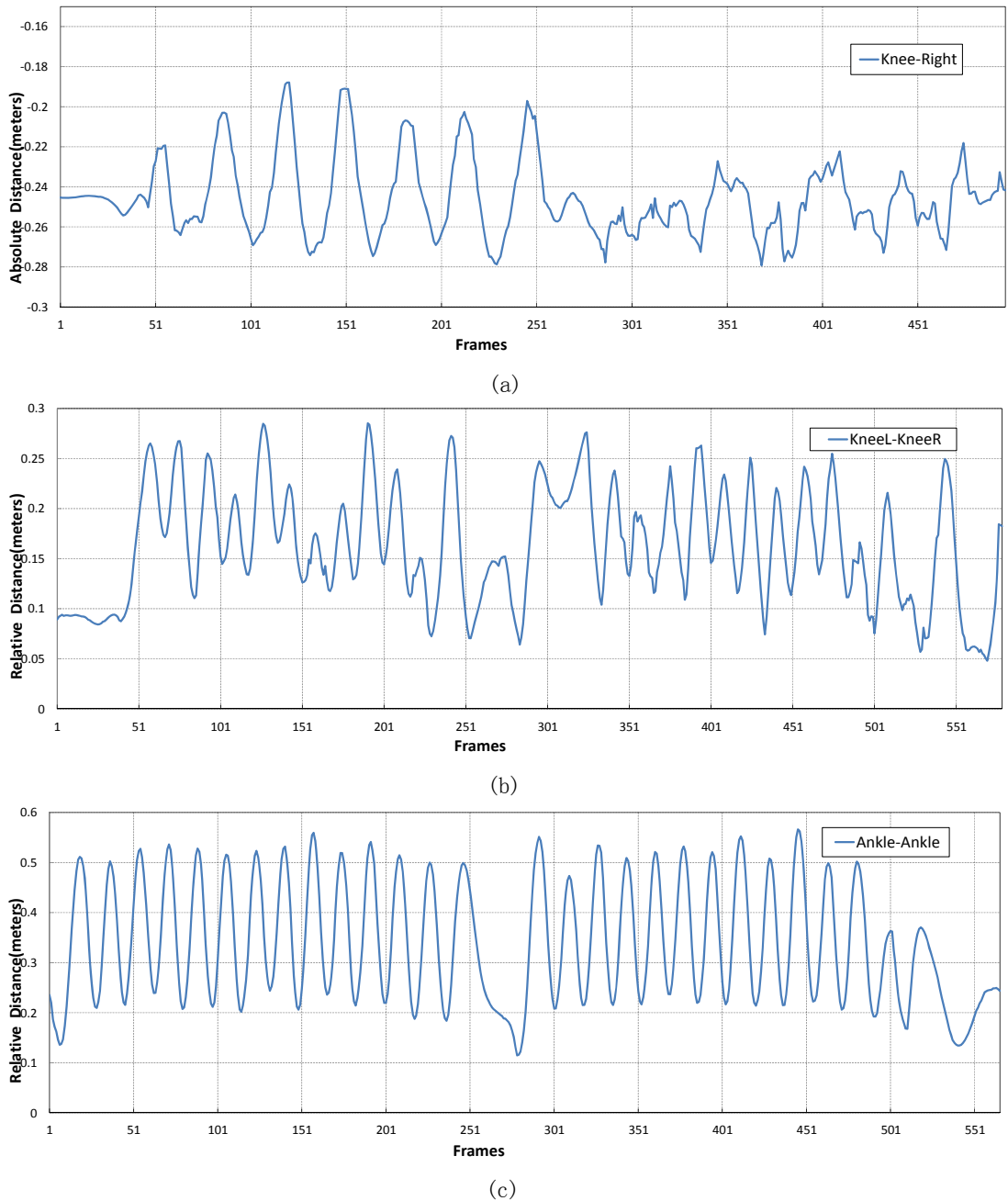

(a)

(b)

(c)

**Fig. 10.** (a) Absolute distance of right knee, in meters, corresponding to a gait sequence (b) relative distance between two knees, in degrees, corresponding to a gait sequence (c) relative distances between two ankles, in meters, corresponding to a gait sequence.

Figure 10 demonstrates that the relative distance features are more robust than the absolute ones. Figure 10(a) shows the absolute coordinate curve of right knee [12] in one gait sequence.

The curve of the second half is in poor condition and almost without patterns. The data capture method shows that the right knee joint is occluded in the second half of one gait sequence. Thus, using these absolute distance features is unreasonable. Figure 10 (b) and (c) give a view of the relative distance features curve between ankles and between knees, respectively. The relative distance curve between ankles is of good quality. The relative distance curve between knees shows a poor quality, but its trend is evident and maintains the gait patterns when observed from the entire sequence. Therefore, the discrimination of the relative distance gait features is higher than angle-based kinematic features and absolute distance features.

## 4. Conclusion

This study investigates the relative distance features for gait recognition with Kinect. Relative distance-based gait features are proposed, where the distances between particular skeleton points and their changes are used to characterize the gait. Random subspace method is also employed for feature selection to further improve the recognition accuracy. The experimental results showed that relative distance features' recognition accuracy reaches up to 85%, which is comparable with the anthropometric features. The two feature sets complement each other well. When relative distance features and anthropometric features are used together, recognition accuracy is at 95%. These results suggest that gait is of enough distinguishing ability. Relative distance features effectively describe gait and are worthy of further study in a non-Kinect scenario.

Future work should investigate relative distance features without Kinect, especially in the scenario where a single RGB camera is used. It is also a possible way to use deep learning method to extract the 2D skeleton model from the RGB image, and then apply our relative distance features method.

**Acknowledgements**

This work was supported by the National Natural Science Foundation of China under Grants 61125201, U1435219, and 61572515.